%% file: main.tex
\documentclass[11pt]{article}

\usepackage[final]{acl}

\usepackage{times}
\usepackage{latexsym}
\usepackage{amssymb}

\usepackage[T1]{fontenc}

\usepackage[utf8]{inputenc}

\usepackage{microtype}

\usepackage{inconsolata}

\usepackage{graphicx}
\usepackage{subcaption}
\usepackage{booktabs}
\usepackage[dvipsnames]{xcolor}
\definecolor{wfmodel}{HTML}{B45309}   
\definecolor{wfcompiler}{HTML}{1D4ED8} 
\definecolor{wfmetric}{HTML}{15803D}  
\usepackage[acronym,nomain,nopostdot,nonumberlist]{glossaries}
\makeglossaries
\glsdisablehyper

\newacronym{llm}{LLM}{large language model}
\newacronym{nlp}{NLP}{natural language processing}
\newacronym{nl}{NL}{natural language}
\newacronym{sql}{SQL}{Structured Query Language}
\newacronym{birdea}{BIRD-EA}{BIRD execution accuracy}
\newacronym{ir}{IR}{Intermediate Representation}

\newcommand{\metricname}{DF-Match}

\newcommand{\capablesize}{68}

\newglossaryentry{multimetric}{
  name={\metricname{} framework},
  description={A question-aware result-set comparison framework that dispatches on two axes (ordering sensitivity and row-count sensitivity) and handles tie and LIMIT boundary-tie conditions}
}
\newglossaryentry{portability}{
  name={portability},
  description={The robustness of a text-to-SQL system across SQL dialects, measured here as the lower-bound / best-dialect accuracy ratio over \{SQLite, PostgreSQL, MySQL, ClickHouse\}: a value near 1.0 (or 100\%) indicates near-identical accuracy across dialects, while a lower value indicates a meaningful drop on at least one dialect}
}
\newglossaryentry{queryplan}{
  name={query plan},
  description={A dialect-agnostic relational algebra operator tree used as the LLM generation target}
}

\title{Closing the Cross-Dialect Gap:\\ Query Plans as a Portable Interface in Text-to-SQL}

\author{
  \textbf{Corentin Royer\textsuperscript{1,2}},
  \textbf{Robin Oester\textsuperscript{1}},
  \textbf{Yotam Perlitz\textsuperscript{1}},
  \textbf{Yannick Metz\textsuperscript{2}},
\\
  \textbf{Andrea Giovannini\textsuperscript{1}},
  \textbf{Mennatallah El-Assady\textsuperscript{2}}
\\
  \textsuperscript{1}IBM Research, Zurich, Switzerland \quad
  \textsuperscript{2}ETH Zurich, Zurich, Switzerland
\\
  \small{
    \textbf{Correspondence:} \href{mailto:corentin.royer@ibm.com}{corentin.royer@ibm.com}
  }
}

\begin{document}
\maketitle
\begin{abstract}
Text-to-SQL systems are typically trained and evaluated on a single dialect (SQLite), yet production deployments span PostgreSQL, MySQL, ClickHouse, and beyond. We show that this single-dialect assumption leads to a substantial drop in lower-bound dialect accuracy for every model we tested. The drop persists across scale, architecture, and even purpose-built text-to-\gls{sql} systems. We argue that the fix is to change the generation target: instead of asking an \gls{llm} to emit dialect-specific \gls{sql}, we have it emit a dialect-agnostic relational algebra \gls{queryplan}, which a deterministic compiler then renders into \gls{sql} for any supported backend. Across thirteen models from 3B to frontier scale, this restores cross-dialect \gls{portability} nearly uniformly, at a small cost in peak accuracy on the model's home dialect for capable prompted models and none once fine-tuned on plans; under matched fine-tuning, plan supervision yields a stronger model than \gls{sql} supervision. We also introduce \metricname{}, a question-aware result-set comparator needed to evaluate fairly across dialects, where existing metrics confound semantic errors with benign cross-dialect variation. More broadly, the result is a reminder that a generation target chosen for \emph{execution} is not necessarily the one that maximizes \emph{generation quality}.
\end{abstract}

%

\section{Introduction}

Natural-language interfaces to databases have advanced rapidly, with \glspl{llm} now achieving over 70\% execution accuracy on standard text-to-SQL benchmarks~\cite{li2025omnisql, gao2024dailsql, talaei2024chess}.
These results suggest that \gls{llm}-based query generation is approaching practical utility.
Yet virtually all existing systems share a critical assumption: they target a single \gls{sql} dialect (typically SQLite), the language of the dominant benchmarks~\cite{yu2018spider, li2024can}.
In production, however, organizations rarely operate a single database engine.
Cloud deployments routinely span PostgreSQL, MySQL, and other backends, thus a text-to-SQL system that works only on SQLite is of limited practical value.

\begin{figure}[t]
  \centering
  \includegraphics[width=\linewidth]{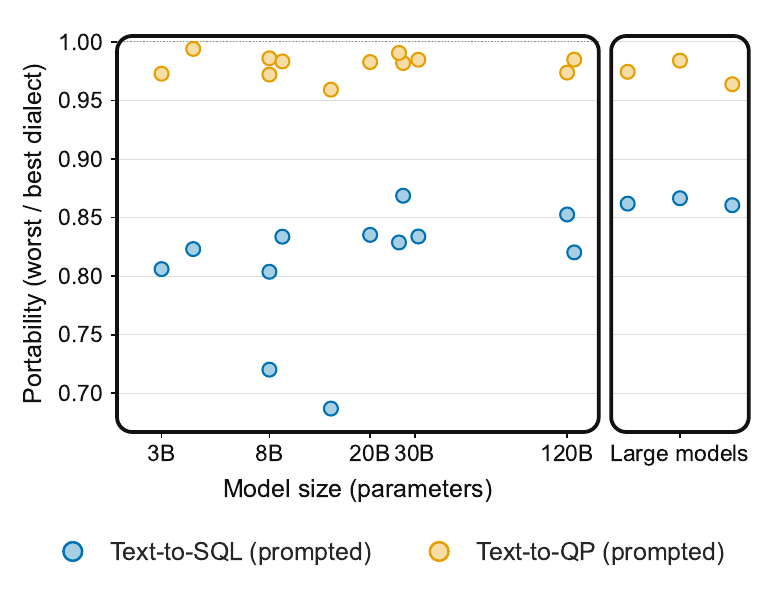}
  \vspace{-15pt}
  \caption{\textbf{Query plans restore cross-dialect portability at every scale.} \Gls{portability} (lower-bound / best-dialect accuracy ratio over \{SQLite, PostgreSQL, MySQL, ClickHouse\}; higher is better) vs.\ model size on \texttt{BIRD}. The \gls{queryplan} cloud sits at $\geq 0.96$ for all sizes; the \gls{sql} cloud is lower and only mildly responsive to scale. Open-weight models on the log-scale left panel; closed-weight frontier models on the right.}
  \label{fig:portability_scatter}
\end{figure}

The cost of this assumption is high. When we evaluate models trained on SQLite across multiple backends, we observe poor \gls{portability} universally: across every model we evaluate, \emph{lower-bound dialect accuracy}, the minimum execution accuracy over the four dialects we evaluate (which is a guarantee on accuracy regardless of the deployed dialect), land 8--17 percentage points below SQLite on \texttt{BIRD}. This degradation is not limited to a single model or architecture; it persists across fine-tuned and prompted models alike, including purpose-built text-to-SQL systems. The root cause is that \gls{sql} dialects differ in function signatures (\texttt{strftime} vs.\ \texttt{to\_char}), type systems (implicit vs.\ explicit casting), regex syntax, and NULL-handling semantics. An \gls{llm} trained to produce SQLite inevitably generates dialect-specific syntax.

We propose \emph{text-to-\gls{queryplan}} as a portable alternative to text-to-\gls{sql}: instead of emitting dialect-specific \gls{sql}, the \gls{llm} generates a single Calcite-compatible~\cite{begoli2018apache} \gls{queryplan} (a tree of operators such as scans, filters, joins, projections, and aggregations, that represents the logical structure of a query without committing to any dialect's syntax), which a deterministic compiler then translates into executable \gls{sql} for the target backend. Because dialect adaptation is handled mechanically rather than by the model, performance across dialects is more reliable. We make the following contributions:

\begin{enumerate}
    \item \textbf{A dialect-agnostic conversion pipeline} between \gls{sql} and Calcite-compatible \glspl{queryplan}, with 94--97\% round-trip fidelity across dialects.

    \item \textbf{\metricname{}}, an ordering- and tie-aware cross-dialect comparator that improves human agreement over \gls{birdea}.

    \item \textbf{An empirical validation of \glspl{queryplan} as a multi-backend generation target}, in four parts:
    \label{sec:intro_preview_anchor}
    \begin{enumerate}
        \item Plans restore \gls{portability} uniformly across scales: the gap in lower-bound dialect accuracy relative to SQLite shrinks from 8--17\,pp to at most 2.3\,pp on \texttt{BIRD} (§\ref{sec:portability}).
        \item For sufficiently capable \glspl{llm} ($\gtrsim\capablesize$B), plans match or exceed \gls{sql} on lower-bound dialect zero-shot accuracy on \texttt{BIRD}, without \gls{queryplan}-specific training (§\ref{sec:claim2_ootb}).
        \item Under matched fine-tuning, plan supervision beats \gls{sql} supervision on accuracy and schema grounding (§\ref{sec:claim3_training}).
        \item Plans show no consistent structural disadvantage: across 13 models, no query-level feature consistently favors one representation (§\ref{sec:claim4_structural}).
    \end{enumerate}
\end{enumerate}

We report lower-bound accuracy throughout, since it is what a practitioner can rely on once the backend is fixed: by that measure the switch is close to free, costing capable prompted models a few points of peak accuracy on their home dialect (§\ref{sec:claim2_ootb}) and nothing once fine-tuned. The takeaway behind these four claims is the same: text-to-\gls{sql} adopted \gls{sql} as a target because \gls{sql} is what databases execute, but a target chosen for \emph{execution} is not necessarily the one that maximizes \emph{generation quality or portability}. Whenever the executable surface (\gls{sql}, in our case) serializes a structured representation (a relational algebra plan), generating the structure and rendering the surface deterministically is a viable alternative.\footnote{Code is available at \url{https://ibm.biz/~tyrUnBWOj}.}

\section{Related Work}
\label{sec:related}

\paragraph{Text-to-SQL and cross-dialect portability.}
Text-to-SQL has shifted from rule-based parsers~\cite{zelle1996learning} and schema-aware encoders such as RAT-SQL~\cite{wang2020rat} to \gls{llm}-based generation, with prompting pipelines like DAIL-SQL~\cite{gao2024dailsql}, DIN-SQL~\cite{pourreza2024dinsql}, and CHESS~\cite{talaei2024chess}, and open-source models trained on large-scale synthetic data~\cite{li2024codes, li2025omnisql}. Nearly all of this work targets a single dialect: SQLite, the language of Spider~\cite{yu2018spider} and BIRD~\cite{li2024can}. PolySQL~\cite{Perlitz2026PolySQLST} first measured the resulting lack of \gls{portability} systematically, uncovering drops in lower-bound dialect accuracy of up to 10\,pp compared to SQLite. Subsequent work targets the same problem but stays at the \gls{sql} level: SQL-GEN~\cite{pourreza2024sql} synthesizes per-dialect training data and merges expert models, MoMQ~\cite{lin2024momq} uses a mixture of dialect experts, and ExeSQL~\cite{zhang2025exesql} bootstraps dialect competence through execution-driven training. These approaches adapt the \emph{model} to each dialect, so the adaptation is repeated for every new dialect and every new model. We instead replace SQL with a dialect-agnostic plan and delegate dialect adaptation to a deterministic compiler, which is adapted once and reused across every model we evaluate.

\paragraph{Intermediate representations as generation targets.}
Generating a structured \gls{ir} rather than surface code has a long lineage, from AST-based semantic parsing~\cite{dong2016language, rabinovich2017asn, yin2018tranx} to constrained decoding over SQL grammars~\cite{scholak2021picard, geng2023grammar}. In the text-to-SQL setting, IRNet's SemQL~\cite{guo2019irnet}, QPL~\cite{eyal2023qpl}, and QueryGym~\cite{ananthakrishnan2026querygym} propose custom intermediates, and Dial~\cite{zhang2026dial} pairs a dialect-aware logical-plan module with execution-driven verification. Closest to our setting, Dialect-SQL~\cite{shi2025dialect} also generates an intermediate representation that is then translated into each target dialect, though it assumes this translation is lossless rather than measuring it. These intermediates are custom DSLs designed for a single pipeline and a single dialect (typically SQLite). The database community has independently developed cross-engine algebra formats with the same dialect-agnostic, algebra-level abstraction: Apache Calcite~\cite{begoli2018apache} and Substrait~\cite{substrait}, a cross-language serialization format designed to decouple query producers from execution engines. Our target is Calcite's algebra \gls{ir}, chosen so that one \gls{queryplan} compiles to multiple production backends. Where prior work shows an \gls{ir} can beat \gls{sql} baselines within one pipeline, we instead ask whether the \gls{ir} is a sound generation target in its own right.

\paragraph{Evaluation metrics.}
Early text-to-SQL benchmarks scored predictions by comparing the SQL itself, e.g. Spider's exact set match over query clauses~\cite{yu2018spider}, which penalizes semantically equivalent rewrites. Executing queries and comparing result sets sidesteps this, and BIRD execution accuracy~\cite{li2024can}, an unordered row-set comparison (ignoring order, duplicates, and cardinality), has become the de facto standard, including on enterprise-backend benchmarks like Spider 2.0~\cite{lei2025spider}. However, Result-based comparison has its own failure modes: false positives in single-instance databases~\cite{zhong2020testsuite}, and structural false negatives whenever ordering or column-shape conventions diverge from the gold query (23\% FP, 29\% FN under ETM~\cite{ascoli2024etm}), both amplified across dialects, where row ordering and type semantics diverge. LLM-based judges~\cite{kim2024flex, zheng2023judging} sidestep these structural mismatches by comparing at a higher level of abstraction, but inherit known biases, including self-preference~\cite{panickssery2024llm}.

\section{Preliminaries}

This section sets up the three concepts this paper builds upon: the text-to-\gls{sql} task, \gls{birdea} as the baseline result-equivalence metric, and the \gls{queryplan} itself as a structured generation target.

\paragraph{Text-to-\gls{sql} Task.}
Let $D$ be a relational database, let $q$ be a natural-language question, and let $p^*$ be a gold SQL query that answers $q$ on $D$. Let $\text{df}(p, D)$ denote the result set obtained by executing query $p$ on $D$, and let $\text{eq}(\cdot, \cdot)$ be a comparison function over result sets (specified below). Given $q$ and $D$, the text-to-SQL task is to generate a query $\hat{p}$ such that $\text{eq}(\text{df}(\hat{p}, D), \text{df}(p^*, D)) = 1$.

\paragraph{BIRD Execution Accuracy.}
A standard choice for $\text{eq}(\cdot,\cdot)$ is \gls{birdea}~\cite{li2024can}, which deems two dataframes equal whenever their rows form the same set, ignoring row order and duplicate counts but remaining sensitive to column shape: $\text{eq}_{\text{BIRD-EA}}(\hat{T}, T^*) = 1$ iff $\text{set}(\hat{T}) = \text{set}(T^*)$. This is the default metric on BIRD and most downstream benchmarks.

\begin{figure*}[t]
  \centering
  \includegraphics[width=\linewidth]{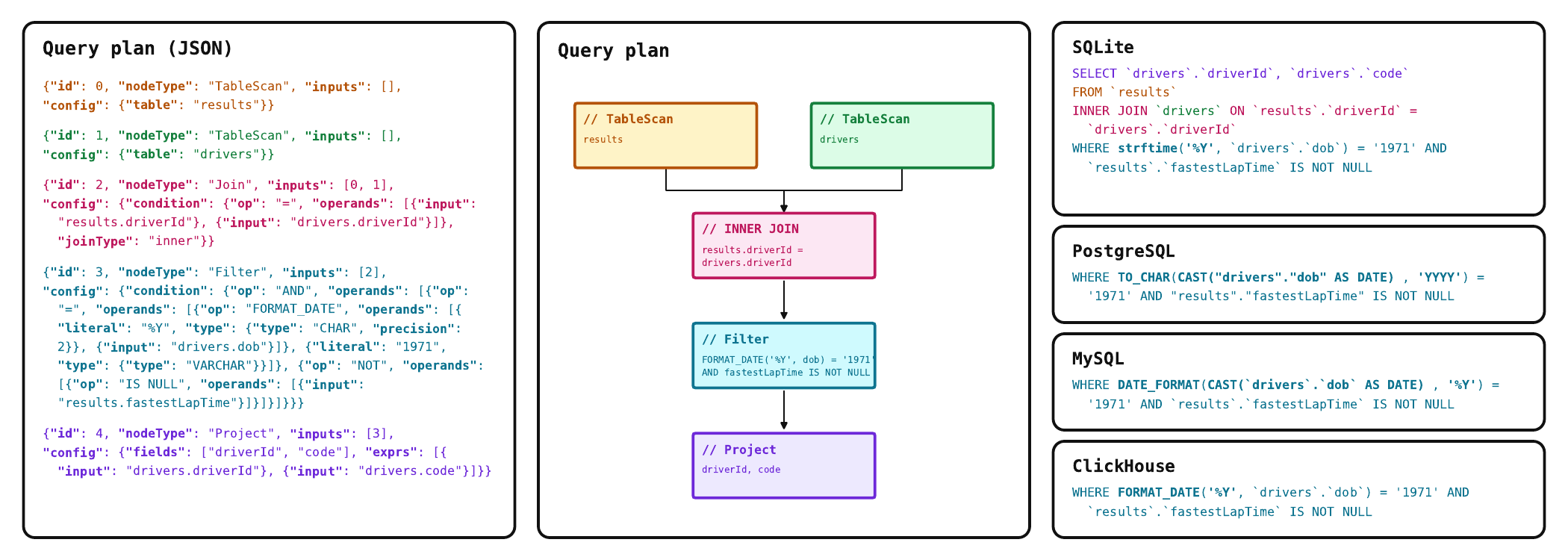}
  \vspace{-20pt}
  \caption{\textbf{Evaluation pipeline, and one query plan in four SQL dialects.} \emph{Top:} the \textcolor{wfmodel}{\gls{llm}} emits a \emph{single} \gls{queryplan}; the \textcolor{wfcompiler}{compiler} renders it to \gls{sql} for every backend; each query executes on its own dialect-parallel copy of the database (§\ref{sec:dual_execution}); and \textcolor{wfmetric}{\metricname{}} compares each predicted dataframe against the gold dataframe from the gold SQLite query. \emph{Bottom:} that same plan shown as emitted JSON (left), as an operator tree (centre), and compiled to the four dialects (right). Colored anchors link each operator to the \gls{sql} fragment it generates. Dialect-specific details, such as function renames (\texttt{strftime}/\texttt{TO\_CHAR}/\texttt{DATE\_FORMAT}) and identifier quoting, sit \emph{inside} the colored spans: the model emits the plan once and never has to choose between them.}
  \label{fig:query_example}
\end{figure*}

\paragraph{Query Plans.}
A \gls{queryplan} $\hat{\pi}$ is an \emph{algebraic} specification of a query: a tree whose nodes are operations on dataframes, where leaves are \texttt{TableScan} nodes that load a base table, and internal nodes (\texttt{Filter}, \texttt{Project}, \texttt{Join}, \texttt{Aggregate}, \texttt{Sort}) transform the dataframes produced by their children (Figure~\ref{fig:query_example}). In contrast to \gls{sql}, which declares \emph{what} result is desired, a plan prescribes \emph{how} it is computed. A deterministic compiler converts $\hat{\pi}$ to an executable query $\hat{p}$ in any supported dialect (Section~\ref{sec:pipeline}).

\section{Method}

Studying \glspl{queryplan} as a multi-backend generation target requires two enablers that the existing text-to-\gls{sql} ecosystem does not provide: a faithful converter between \gls{sql} and \glspl{queryplan}, so that plans can be used as a generation target (§\ref{sec:pipeline}); and a result-set metric that tolerates the cross-dialect variation that yield errors with \gls{birdea}, so that any generation target can be evaluated fairly across backends (§\ref{sec:multimetric}). Figure~\ref{fig:query_example} shows how the two fit together across the four stages of the workflow, and in particular where the conversion pipeline is used in each direction.

\subsection{Conversion pipeline}
\label{sec:pipeline}

The pipeline runs in both directions. The \emph{forward} direction (SQL $\to$ plan) converts gold SQLite queries into reference plans $\pi^*$. The \emph{backward} direction (plan $\to$ SQL) compiles any plan $\hat{\pi}$ into executable SQL for a target backend $E$. Both directions are built on Apache Calcite~\cite{begoli2018apache}: the \gls{ir} is a JSON serialization of a Calcite logical query plan, which the forward direction produces and the backward direction consumes.

We measure the fidelity of the pipeline by \emph{round-trip equivalence}: starting from a SQL query $p$, the forward direction produces a plan $\pi$, and the backward direction recompiles $\pi$ to a SQL query $\hat{p}_E$ on backend $E$. The round-trip succeeds when the recompiled query yields the same result as the original, i.e.\ $\text{eq}(\text{df}(p, D), \text{df}(\hat{p}_E, D)) = 1$. Section~\ref{sec:pipeline_reliability} reports this rate per backend on \texttt{BIRD} and \texttt{Spider}.

\paragraph{Forward: SQL $\to$ plan.} Benchmarks ship gold queries in SQLite, but Calcite's SQL parser does not accept SQLite syntax. We therefore convert each gold query $p^*$ in three steps:

\noindent\textbf{(1)} \emph{SQLite $\to$ MySQL transpile} using SQLGlot~\cite{sqlglot}. We target MySQL because its syntax is closest to SQLite, minimizing the transpilation errors.

\noindent\textbf{(2)} \emph{MySQL $\to$ Calcite plan}: Calcite parses the rewritten query into a logical operator tree. We disable Calcite's optimization rules so the plan shape mirrors the original query; otherwise, rewrites such as predicate pushdown, or stripping \texttt{IS NOT NULL} checks on columns Calcite considers non-nullable, would alter what the \gls{llm} has to learn and break portability to backends with different nullability semantics.

\noindent\textbf{(3)} \emph{Serialize} the tree to a compact JSON format (one object per operator; Figure~\ref{fig:query_example}) with symbolic column references so the encoding is robust across reorderings, yielding $\pi^*$.

\paragraph{Backward: plan $\to$ SQL.} Given a plan $\pi$ and a target backend $E \in \{\text{SQLite}, \text{PostgreSQL}, \text{MySQL}, \text{ClickHouse}\}$, we deserialize $\pi$ into a Calcite logical plan and emit SQL through a \emph{custom Calcite dialect} we implement for each backend. Our dialects perform two kinds of adaptation. \emph{Syntactic}: identifier quoting, \texttt{LIMIT}/\texttt{OFFSET} layout, function renaming such as \texttt{FORMAT\_DATE} $\to$ \texttt{STRFTIME}/\texttt{TO\_CHAR}/\texttt{DATE\_FORMAT}. \emph{Semantic}: aligning behavior across backends. Examples include: enforcing SQLite's NULL-first sort order when emitting for PostgreSQL; rewriting \texttt{LIKE} to \texttt{ILIKE} to preserve SQLite's case-insensitive match on PostgreSQL; wrapping string-to-integer casts to mimic SQLite's lenient conversion where PostgreSQL would raise; and keeping textual comparisons textual instead of letting PostgreSQL's stricter typing coerce them to numeric. 


Concretely, instantiating round-trip equivalence with $\text{eq} = \text{eq}_{\text{BIRD-EA}}$ yields a large number of spurious failures driven by cross-dialect row ordering and tie-breaking rather than real semantic drift, motivating the ordering- and tie-aware comparator we introduce next.

\subsection{The \metricname{} framework}
\label{sec:multimetric}

\gls{birdea} has two main failure modes even in single-dialect evaluation, with high false-positive and false-negative rates~\cite{ascoli2024etm, kim2024flex}; cross-dialect evaluation compounds both, as we show empirically in §\ref{sec:metric_validation}. Consider two questions where \gls{birdea} fails in opposite directions. \emph{(a)~``List the top three clients by spending, in order.''} A model that returns the correct three clients in the wrong order is accepted by \gls{birdea} because \texttt{set(predicted) == set(gold)} ignores row order (\emph{false positive}). \emph{(b)~``Which client spent the most?''} with gold query \texttt{ORDER BY total DESC LIMIT 1}. If two clients are tied at the top, the gold reference contains whichever one the backend returned first; a model that returns the other tied client is rejected by \gls{birdea} (\emph{false negative}).

\metricname{} addresses both by deciding, \emph{per question}, what counts as a correct result. It (i) classifies each question along two axes that determine which constraints to enforce, (ii) extracts tie information from the gold SQL so that semantically equivalent rows can be permuted, and (iii) compares the predicted and gold dataframes under per-class rules built from the previous two ingredients.

\paragraph{Question classification.}
\label{sec:multimetric:classification}
Each question is labeled along two axes: \textbf{ordering sensitivity}, whether the question asks for a ranked result (``rank\ldots'', ``top five\ldots''); and \textbf{row-count sensitivity}, whether it pins the output cardinality (``the highest-spending client'' $\to$ 1). We classify the \emph{natural-language question}, not the gold SQL, because the gold query is sometimes under-specified whereas the question carries the intent. The two booleans are produced by an LLM classifier (validated in §\ref{sec:metric_validation}; full protocol in Appendix~\ref{sec:appendix_metric_validation}). Four combinations are possible, but the \texttt{unordered-fixed} cell is empty in practice, leaving three classes: \texttt{not\_ordered}, \texttt{ordered\_fixed}, and \texttt{ordered\_not\_fixed}.

\paragraph{Tie handling.}
\label{sec:multimetric:ties}
Once a question is identified as ordered, the comparison rules need to know which result rows are \emph{tied} under the gold ordering before permitting legitimate reorderings, and which rows the backend could legitimately have placed at a \texttt{LIMIT} boundary. \metricname{} derives both by AST-rewriting the gold SQL into two auxiliary queries: the \emph{order-key query}, which recovers the sort values per row so that consecutive equal values define a \emph{tie group}, and the \emph{boundary-alternatives query}, which enumerates rows the gold could have placed at a \texttt{LIMIT} cut. Both are executed once per gold query and cached; the next paragraph shows them in use.

\paragraph{Worked example.}
Returning to the two questions from the top of this section. Question (a), ``List the top three clients by spending, in order,'' is classified \texttt{ordered\_fixed}. From a gold query \texttt{SELECT name FROM clients ORDER BY total\_spend DESC LIMIT 3}, the order-key query \texttt{SELECT total\_spend FROM clients ORDER BY total\_spend DESC} returns the sort values; consecutive equal values define tie groups. The comparison enforces positional row-by-row equality, but accepts any permutation of rows whose positions fall in the same tie group: a prediction that swaps two clients tied at the same \texttt{total\_spend} is accepted; a prediction that swaps two clients with distinct spends is rejected. Question (b), ``Which client spent the most?'' (gold \texttt{ORDER BY total\_spend DESC LIMIT 1}), is classified \texttt{ordered\_not\_fixed}. With $K=1$ and a tie at the top (say at \texttt{total\_spend = 90}), the boundary-alternatives query becomes \texttt{SELECT name FROM clients WHERE total\_spend = 90}, enumerating every client the backend could have placed at the cut; any of them is accepted, addressing the \gls{birdea} false negative.

\paragraph{Per-class comparison.}
\label{sec:multimetric:compare}
Comparing predicted $\hat{T}$ to gold $T^*$ first maps gold columns to predicted columns by matching distinct value sets (tolerating duplicate-count differences from \texttt{DISTINCT}/\texttt{GROUP BY} divergences); a gold column with no match rejects $\hat{T}$. The class-specific row check then proceeds:
\begin{itemize}
    \item \texttt{not\_ordered}: row multisets must match, tolerating duplicate-count differences from \texttt{DISTINCT}/\texttt{GROUP BY} divergences.
    \item \texttt{ordered\_fixed}: cardinalities must be equal and rows must match positionally, modulo permutations within tie groups and consecutive duplicates.
    \item \texttt{ordered\_not\_fixed}: $|\hat{T}| \ge |T^*|$, the predicted prefix must match the gold as in \texttt{ordered\_fixed}, with boundary alternatives accepted when the cut falls inside a tie group.
\end{itemize}
Numeric comparisons use $\varepsilon=10^{-9}$ throughout. Pseudocode for the per-class dispatch is in Appendix~\ref{sec:metric_pseudocode}.

\paragraph{Relation to \gls{birdea}.}
\label{sec:multimetric:bird}
\gls{birdea} is a permissive instance of the \texttt{not\_ordered} branch: it compares tuple sets, so row order, duplicate counts, and cardinality are all ignored. This exposes it to both false positives and false negatives. \metricname{} is designed so that, given a correct question classification (validated in §\ref{sec:metric_validation}), it never produces false positives (sketch in Appendix~\ref{sec:metric_conservatism}): its accuracy is therefore a \emph{lower bound} on true accuracy, whereas \gls{birdea} is neither a lower nor an upper bound.

\section{Experimental setup}

We evaluate \gls{queryplan} generation against direct SQL generation across four dialects, using a diverse set of models and a controlled fine-tuning comparison. This section describes the benchmarks (§\ref{sec:benchmarks}), the cross-dialect evaluation environment (§\ref{sec:dual_execution}), the models we compare (§\ref{sec:models}), our training setup (§\ref{sec:training}), and the error-categorization procedure used in our analysis (§\ref{sec:error_categorization}).

\subsection{Benchmarks \& Datasets}
\label{sec:benchmarks}

We evaluate on the dev splits of two standard text-to-SQL benchmarks: \texttt{BIRD}~\cite{li2024can} and \texttt{Spider}~\cite{yu2018spider}. Both ship with SQLite databases, gold SQLite queries, and natural-language questions spanning multiple domains. We use the dev splits because the test sets are not publicly available. For fine-tuning (§\ref{sec:training}), we additionally draw training data from \texttt{SynSQL}~\cite{li2025omnisql}, a large-scale synthetic text-to-SQL corpus. \texttt{SynSQL} is used only as a training source and is not part of our evaluation. Licenses and intended-use details for all three datasets are in Appendix~\ref{sec:appendix_datasets}.

\subsection{Dual Execution}
\label{sec:dual_execution}

Cross-dialect evaluation requires the same question executed against multiple backends. We follow \citet{Perlitz2026PolySQLST} and migrate every SQLite database in \texttt{BIRD} and \texttt{Spider} to PostgreSQL, MySQL, and ClickHouse via schema-level ETL using the DLT data-loading library~\cite{dlt}, yielding four dialect-parallel copies with identical schemas, contents, and questions. Each predicted query runs against the corresponding backend and is evaluated with \metricname{} (§\ref{sec:multimetric}).

\subsection{Models \& Baselines}
\label{sec:models}

Each model is evaluated in two regimes. The \gls{sql} regime prompts the model once per target dialect (four runs over SQLite, PostgreSQL, MySQL, ClickHouse) with a one-shot example differing only in the target-dialect name. The \gls{queryplan} regime prompts the model once with a one-shot example plus a brief specification of the plan language (operator set and syntax), and our pipeline (§\ref{sec:pipeline}) compiles the emitted plan to executable \gls{sql} for all four dialects. The model set spans three size tiers and mixes open-weights, API-served, dense, and mixture-of-experts architectures (full list with parameter counts in Table~\ref{tab:portability_results}); we additionally include two purpose-built text-to-\gls{sql} systems, OmniSQL~\cite{li2025omnisql} and SQLCoder\footnote{\url{https://huggingface.co/defog/sqlcoder-7b-2}}, as a reference point for how SQLite-specialized models generalize off-dialect. All prompted runs use greedy decoding.

\subsection{Training}
\label{sec:training}

We fine-tune \texttt{Ministral-3-8B} on a 1M-example mixture drawn from \texttt{SynSQL}~\cite{li2025omnisql} and the \texttt{BIRD-train} split, comparing two supervision targets: the gold SQL query $p^*$, or a gold plan $\pi^*$ obtained by passing $p^*$ through the SQL$\to$Calcite pipeline (§\ref{sec:pipeline}). Examples where the pipeline fails are excluded from \emph{both} runs, so the \gls{sql} and \gls{queryplan} fine-tunes see identical questions and differ only in the generation target. We train with LoRA~\cite{hu2022lora}; optimizer, learning-rate schedule, and compute budget are held constant across runs. Full hyperparameters are in Appendix~\ref{sec:training_appendix}.

\subsection{Error Categorization}
\label{sec:error_categorization}

To understand what kinds of errors each generation target produces, we classify each error with an LLM judge (\textit{Claude Haiku 4.5}, temperature~0) into one of six categories: Schema Linking, Filtering/Logic, Aggregation/Grouping, Ordering/Ranking, Dialect Syntax, or Other. The judge sees the question, gold and predicted SQLite queries, and any execution error; \glspl{queryplan} are compiled to SQLite first so the judge sees the same input distribution regardless of generation target. Category definitions and the user-prompt template are in Appendix~\ref{sec:llm_judge_prompt}. Hand-checking a sample of its labels against manual annotation gives 76\% precision across the six categories. The judge only sorts errors that have already been detected by \metricname{} into buckets, so its accuracy shapes the breakdown in §\ref{sec:claim4_structural} but does not enter any headline metric.

\section{Results}

We organize this section to mirror the contributions: §\ref{sec:pipeline_reliability} and §\ref{sec:metric_validation} first validate the two instruments (the conversion pipeline and \metricname{}), §\ref{sec:claim1_gap}--§\ref{sec:claim4_structural} then address the four claims about \glspl{queryplan} as a generation target.

\subsection{The conversion pipeline is reliable}
\label{sec:pipeline_reliability}

Table~\ref{tab:pipeline_reliability} reports round-trip accuracy of the conversion pipeline under \metricname{} on \texttt{BIRD} and \texttt{Spider} across the four backends. The pipeline achieves 94--97\% fidelity everywhere. The remaining 3--6\% are dominated by errors in the gold query itself and by cross-backend typing mismatches that surface only after compilation (for example, implicit string-to-numeric coercions that SQLite tolerates but PostgreSQL or ClickHouse reject). These queries are retained at evaluation time: a round-trip failure on the gold query does not exclude a model from generating a different but semantically correct \gls{queryplan} that answers the question, so excluding them would bias the benchmark toward plan-convertible questions.

\begin{table}[t]
  \centering
  \small
  \setlength{\tabcolsep}{4pt}
  \begin{tabular}{l rrrr}
    \toprule
    & \textbf{SQLite} & \textbf{PG} & \textbf{MySQL} & \textbf{CH} \\
    \midrule
    \texttt{BIRD}   & 97.1 & 95.4 & 95.6 & 94.7 \\
    \texttt{Spider} & 97.2 & 96.7 & 96.6 & 95.6 \\
    \bottomrule
  \end{tabular}
  \caption{Round-trip conversion accuracy (\%) of the pipeline under
  \metricname{}. PG = PostgreSQL, CH = ClickHouse. Each gold SQLite query is
  converted to a Calcite plan and back to the target dialect; accuracy measures
  whether the reconstructed query returns an equivalent result as the original.}
  \label{tab:pipeline_reliability}
\end{table}

\subsection{\metricname{} agrees with human judgment}
\label{sec:metric_validation}

Before reporting results under \metricname{}, we validate its performance.
The LLM classifier that drives \metricname{}'s per-class comparison reaches
\textbf{93.4\%} balanced accuracy (mean of per-class recall) on the ordering-sensitivity axis and
\textbf{99.5\%} on the row-count-sensitivity axis, measured on 103
hand-annotated \texttt{BIRD} questions. Applied to the full
\texttt{BIRD} set, the classifier labels 86.6\% of questions as
\texttt{not\_ordered}, 11.9\% as \texttt{ordered\_fixed}, and 1.5\% as
\texttt{ordered\_not\_fixed}. On a separate sample of 103 cases where \metricname{} and
\gls{birdea} return opposite verdicts, a human annotator sides with
\metricname{} in 96.1\% of cases. Replacing \gls{birdea} with \metricname{}
also raises apparent round-trip pipeline reliability by 14--16 pp per
dialect. Full protocols, per-axis numbers, the
\gls{birdea} comparison table, and study limitations are in
Appendix~\ref{sec:appendix_metric_validation}.

\subsection{\Glspl{queryplan} restore \gls{portability}}
\label{sec:claim1_gap}
\label{sec:portability}

Figure~\ref{fig:portability_scatter} plots \gls{portability} on \texttt{BIRD} against model size for every model we evaluate. The pattern is the same at every scale and architecture: \gls{sql} targets land at 0.69--0.87 \gls{portability} with lower-bound accuracy drops of 8--17\,pp, while \gls{queryplan} targets land at $\geq 0.96$ with lower-bound accuracy drops at most 2.3\,pp. Neither cloud trends with model size: scaling does not buy \gls{sql} portability. The drop is driven by ClickHouse and PostgreSQL; MySQL accuracy tracks SQLite for most models. Per-dialect numbers for every model are in Appendix~\ref{sec:appendix} (Table~\ref{tab:portability_results}), and the same trend holds on Spider (Appendix~\ref{sec:appendix_spider}). Whether this comes at a cost in peak accuracy on the model's home dialect is the question we address next.

\subsection{Capable models produce \glspl{queryplan} zero-shot}
\label{sec:claim2_ootb}

\begin{figure}[t]
  \centering
  \includegraphics[width=\linewidth]{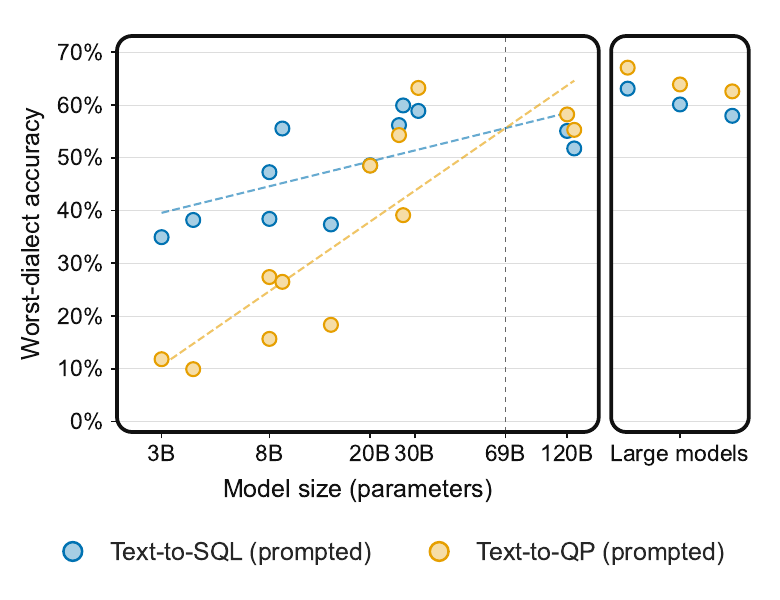}
  \vspace{-15pt}
  \caption{\textbf{Plans win above $\sim\capablesize$B.} Lower-bound dialect accuracy on \texttt{BIRD} vs.\ model size: per model, the lowest accuracy across the four dialects. The \gls{queryplan} trend rises more steeply than \gls{sql}; the two cross at $\sim\capablesize$B (dashed line), beyond which plans deliver as much or more lower-bound dialect accuracy than \gls{sql}. Per-model breakdown in Table~\ref{tab:portability_results}.}
  \label{fig:capability_threshold}
\end{figure}

Figure~\ref{fig:capability_threshold} reports lower-bound dialect accuracy on \texttt{BIRD} as a function of model size. The \gls{queryplan} trend rises more steeply with scale than the \gls{sql} trend, and the two cross around \capablesize{}B parameters. We therefore use \emph{capable} operationally to mean prompted models above approximately \capablesize{}B parameters: above this threshold, \glspl{queryplan} deliver at least as much lower-bound dialect accuracy as \gls{sql} on \texttt{BIRD}, with no \gls{queryplan}-specific training and from a single format demonstration. On \texttt{Spider}, where the \gls{sql} baseline is already close to its ceiling, the two targets are instead near parity for these models. Below it, the \gls{queryplan} regime is strictly harder for the model: across the small and mid prompted band, plan-format errors drag lower-bound dialect accuracy 19--29\,pp below the \gls{sql} baseline, and the cross-over has not yet occurred.


\subsection{\Glspl{queryplan} are a strong training target}
\label{sec:claim3_training}
\label{sec:error_analysis}

\begin{figure}[t]
  \centering
  \includegraphics[width=\linewidth]{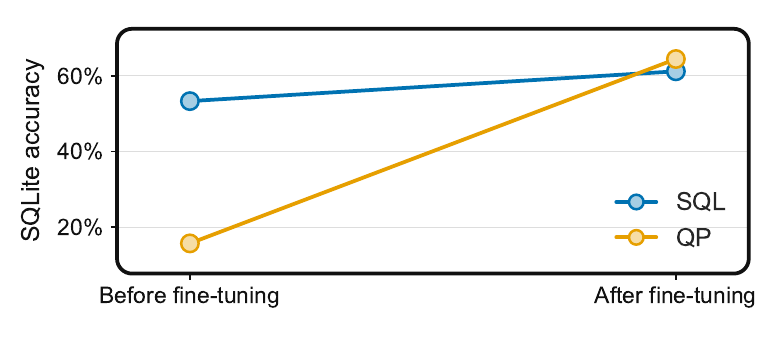}
  \vspace{-15pt}
  \caption{\textbf{At matched cost, plans yield a stronger model.} SQLite accuracy on \texttt{BIRD} for Ministral-3-8B before and after fine-tuning, on identical data. Prompted \gls{sql} starts well above prompted \gls{queryplan}, but \gls{queryplan} supervision overtakes \gls{sql} supervision after fine-tuning.}
  \label{fig:finetune_inversion}
\end{figure}

Holding training data, optimizer, and compute identical (§\ref{sec:training}), we fine-tune Ministral-3-8B twice, once on gold \gls{sql}, once on gold \glspl{queryplan}, and evaluate both on \texttt{BIRD}. The \gls{queryplan}-supervised model reaches \textbf{64.5\%} on SQLite versus \textbf{61.2\%} for the \gls{sql}-supervised model, while keeping the lower-bound accuracy drop at 1.7~pp (portability 0.97 vs.\ 0.82 for the \gls{sql} fine-tune; Table~\ref{tab:portability_results}). \Glspl{queryplan} therefore do more than inherit the \gls{sql} baseline: at matched cost, they produce a \emph{better} model. This is despite a substantially lower starting point: prompted Ministral-3-8B reaches 53.3\% under \gls{sql} but only 15.7\% under \glspl{queryplan}, so \gls{queryplan} supervision lifts the model by 48.8~pp against the 7.9~pp gain from \gls{sql} supervision. 


\subsection{\Glspl{queryplan} have no consistent structural disadvantage}
\label{sec:claim4_structural}
\label{sec:interpretability}

If \glspl{queryplan} carried an intrinsic cost, it would surface either as a \emph{category} of errors elevated under \glspl{queryplan}, or as a query-level \emph{structural feature} that predicts \gls{queryplan} losses. We check both and find no such signal. 

\begin{figure}[t]
  \centering
  \includegraphics[width=\linewidth]{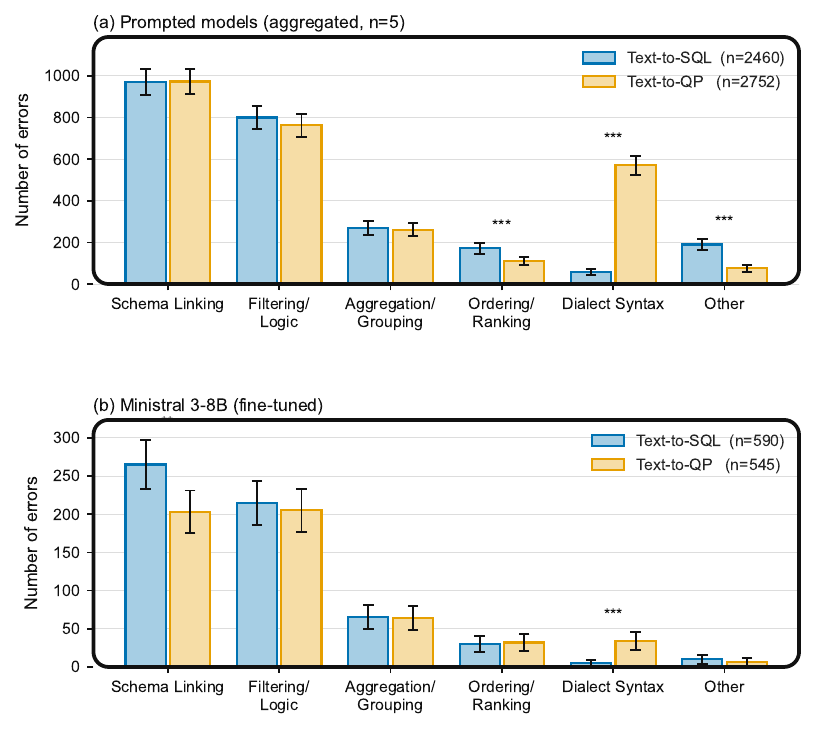}
  \vspace{-15pt}
  \caption{\textbf{No reasoning-error category penalizes plans.} LLM-classified error counts on \texttt{BIRD}. (a) the 5 strongest prompted models aggregated (3 frontier models, GPT-OSS 120B, Mistral Medium); (b) fine-tuned Ministral-3-8B. Reasoning categories are near-identical between \gls{sql} and \gls{queryplan}. The only \gls{queryplan} surplus is Dialect Syntax (panel a), which vanishes after fine-tuning (panel b).}
  \label{fig:error_combined}
\end{figure}

\paragraph{Reasoning errors are tied; the surplus is a decoding tax.}
Across the 5 strongest prompted models (Figure~\ref{fig:error_combined}a), the reasoning categories (Schema Linking, Filtering/Logic, Aggregation/Grouping) are essentially matched between \gls{sql} and \gls{queryplan}. The only \gls{queryplan} surplus concentrates in Dialect Syntax, and inspection shows these are overwhelmingly malformed-JSON outputs: a decoding-level tax that constrained decoding or resampling could plausibly address, and one that largely collapses after LoRA fine-tuning at 8B (Figure~\ref{fig:error_combined}b). The surplus is also a property of plan \emph{generation}, not of compilation: separating failures that occur on all four backends (a wrong plan) from those on a single backend (a compilation defect), compilation accounts for only 2--3\% of a dialect's failures on \texttt{BIRD} and 5--7\% on \texttt{Spider}, consistent with the round-trip fidelity on gold plans in §\ref{sec:pipeline} (94--97\%).

\paragraph{Sort and limit errors \emph{drop} under \glspl{queryplan}.}
A small structural \emph{advantage} of the \gls{queryplan} format is that ordering and cardinality are explicit operators (\textsc{Sort}, \textsc{Aggregate}) rather than parsed clauses. Empirically, Ordering/Ranking errors drop from 172 under \gls{sql} to 109 under \gls{queryplan} on the same 5 models, a 37\% reduction in the one category where the surface forms differ most.

\paragraph{No query-level feature consistently favors one representation.}
A structural cost might also hide in queries with particular shapes without producing a distinct error type. We test for this with a per-model elastic-net analysis over 21 query-level features across 13 models (Appendix~\ref{sec:appendix_features}); the result is null. No feature has a stable signed effect across models, and the one feature with a non-zero pooled coefficient (\verb|ps_has_scalar_subquery|) shows substantial between-model heterogeneity ($I^2=66\%$). Per-model preferences exist but disagree, consistent with model-internal factors rather than a shared structural cause.

\section{Conclusion}

We argued that \gls{sql}'s role as the default generation target for text-to-database systems is a legacy assumption worth revisiting. Generating a dialect-agnostic Calcite plan and letting a deterministic compiler handle dialect adaptation restores cross-dialect \gls{portability}, shrinking the lower-bound accuracy drop from 8--17\,pp under \gls{sql} to at most 2.3\,pp across every model we evaluated, including a LoRA-tuned Ministral-3-8B. We also introduced \metricname{}, a question-aware result-set comparator that fixes the cross-dialect failure modes of \gls{birdea} and is needed to evaluate any text-to-\gls{sql} system fairly across backends.

\paragraph{Scope and outlook.}
Our evaluation covers four open-source backends (SQLite, PostgreSQL, MySQL, ClickHouse); extending the compiler to Snowflake or BigQuery, whose operator surfaces diverge further from Calcite's core, is a natural next step, as is a broader sweep across model families and sizes. Beyond portability, the \gls{queryplan} format opens several axes worth studying: \emph{extensibility}, since Calcite's operator algebra accepts new operators (e.g., semantic operations such as embedding-based retrieval or LLM-judged predicates) without changing the generation interface; \emph{constrained decoding} to eliminate the residual malformed-JSON failures; \emph{test-time scaling} via verifier-guided sampling over a typed operator tree; and \emph{interpretability}, since an operator tree is directly visualizable. We leave these to future work.

\section*{Limitations}

We discuss limitations along four axes.

\paragraph{Backend coverage.}
Our compiler and evaluation cover four open-source backends (SQLite, PostgreSQL, MySQL, ClickHouse). Cloud warehouses such as Snowflake and BigQuery, whose operator surfaces diverge further from SQLite's syntax, are not evaluated.

\paragraph{Benchmark coverage.}
We evaluate on the dev splits of \texttt{BIRD} and \texttt{Spider}, both of which ship gold queries in SQLite and are English-only. Hidden test sets are not publicly accessible, so we cannot rule out dev-set contamination in the reported numbers. Cross-lingual text-to-SQL is out of scope.

\paragraph{\metricname{} validity.}
The conservatism argument in Appendix~\ref{sec:metric_conservatism} assumes a correct question classification; misclassifying an ordered question as \texttt{not\_ordered} weakens the comparison to a set check and can introduce false positives, with the empirical rate bounded by the classifier accuracy reported in §\ref{sec:metric_validation}. The human-agreement study in Appendix~\ref{sec:appendix_metric_validation} is single-labeler and drawn from the disagreement set rather than i.i.d.\ from \texttt{BIRD}, so the reported 96.1\% should be read as a directional rather than population estimate.

\paragraph{Best-dialect accuracy.}
Restoring portability is not free at the top of the SQLite-only leaderboard. For most prompted models, \glspl{queryplan} land a few points below \gls{sql} on the model's home dialect (Table~\ref{tab:portability_results}); the gap closes for capable models above $\sim\capablesize$B and inverts under fine-tuning at 8B (§\ref{sec:claim2_ootb}, §\ref{sec:claim3_training}). A practitioner whose deployment targets only SQLite, and who is willing to accept the cross-dialect fragility, can still extract slightly higher peak accuracy from direct \gls{sql} generation with a sub-capability-threshold model.

\paragraph{Mitigations not evaluated.}
We hypothesize but do not measure that constrained decoding or resampling would close the residual malformed-JSON surplus reported in §\ref{sec:claim4_structural}; we leave this to future work along with the broader directions discussed in the conclusion.

\bibliography{sample}

\printglossaries

\clearpage

\appendix

\section{Full Results Table}
\label{sec:appendix}

Tables~\ref{tab:portability_results} and~\ref{tab:portability_results_spider} report per-dialect execution accuracy for
all models on \texttt{BIRD} and \texttt{Spider} respectively.
The lower-bound accuracy drop for SQL models is driven by ClickHouse and PostgreSQL,
while MySQL accuracy remains close to SQLite across all models.
\Gls{queryplan} generation shrinks the drop to at most 2.3\,pp
(\gls{portability} $\geq 0.96$) across every model, including fine-tuned Ministral-3-8B.

Both tables also report the mean accuracy across the four dialects and, on each model's
\gls{queryplan} row, the \gls{queryplan}$-$\gls{sql} difference in best-dialect ($\Delta$Peak) and
worst-dialect ($\Delta$Worst) accuracy, so that the peak-accuracy cost of the switch is legible next
to the \gls{portability} gain rather than hidden behind a ratio.

\input{figures/fig_portability_table}

\section{Spider Figures}
\label{sec:appendix_spider}

\begin{figure}[h]
  \centering
  \includegraphics[width=\linewidth]{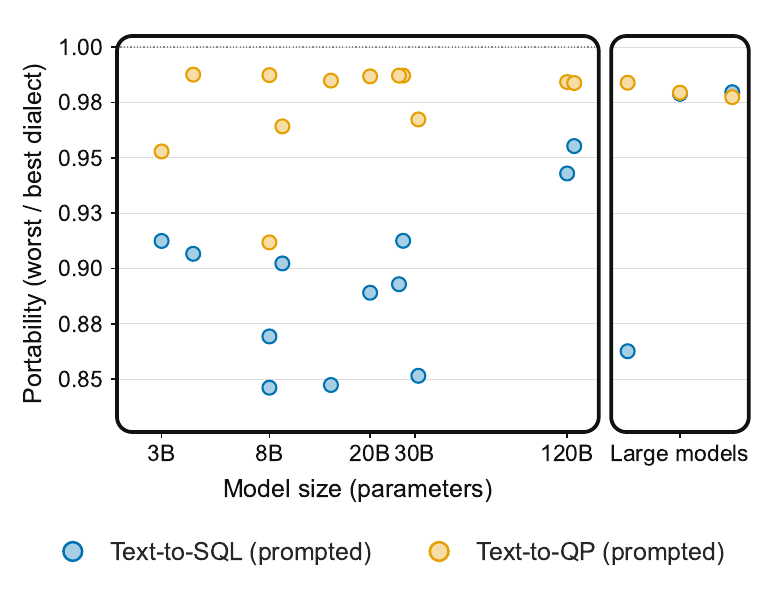}
  \vspace{-15pt}
  \caption{\Gls{portability} (lower-bound / best-dialect accuracy ratio) vs.\ model size on \texttt{Spider}. Same encoding as Figure~\ref{fig:portability_scatter}.}
  \label{fig:portability_scatter_spider}
\end{figure}

\begin{figure}[h]
  \centering
  \includegraphics[width=\linewidth]{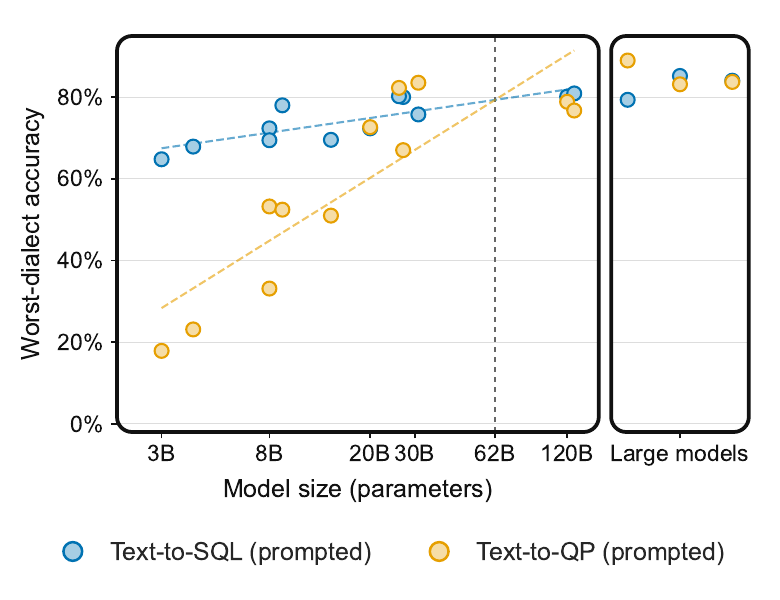}
  \vspace{-15pt}
  \caption{Lower-bound dialect accuracy on \texttt{Spider} vs.\ model size. Same encoding as Figure~\ref{fig:capability_threshold}.}
  \label{fig:capability_threshold_spider}
\end{figure}

The same trends seen on \texttt{BIRD} hold on \texttt{Spider}. Figure~\ref{fig:portability_scatter_spider} shows \gls{portability} vs.\ model size: the \gls{queryplan} cloud sits above the \gls{sql} cloud at every scale, with neither cloud trending with size. Figure~\ref{fig:capability_threshold_spider} shows lower-bound dialect accuracy: the \gls{queryplan} trend rises with scale and crosses \gls{sql} near 62B, slightly below the \texttt{BIRD} crossover at \capablesize{}B but consistent with the same overall picture.

\section{Training Details}
\label{sec:training_appendix}

We fine-tune Ministral-3-8B on 1M examples drawn from \texttt{SynSQL} and \texttt{BIRD-train} (§\ref{sec:training}). The \gls{sql} and \gls{queryplan} runs use identical data, optimizer, schedule, and compute budget; the only difference is the supervision target.

\paragraph{LoRA configuration.}
\begin{itemize}
  \item Rank: 64
  \item $\alpha$: 16 (fixed; \emph{not} $2\cdot\text{rank}$)
  \item Dropout: 0.05
  \item Target modules: all linear layers
  \item Trainable parameters: $\approx$200M
\end{itemize}

\paragraph{Optimization.}
\begin{itemize}
  \item Optimizer: AdamW
  \item Learning rate: $1\times 10^{-4}$
  \item Schedule: cosine decay with linear warmup
  \item Epochs: 1
  \item Batch size: 64
\end{itemize}

\paragraph{Software.} Fine-tuning uses Hugging Face \texttt{transformers}\footnote{\url{https://github.com/huggingface/transformers}} with \texttt{peft}\footnote{\url{https://github.com/huggingface/peft}} for the LoRA adapters~\cite{hu2022lora}. Inference for all open-weight models is served with vLLM\footnote{\url{https://github.com/vllm-project/vllm}}. The conversion pipeline uses Apache Calcite~\cite{calcite} and SQLGlot~\cite{sqlglot} (§\ref{sec:pipeline}), and the cross-dialect database migration uses DLT~\cite{dlt} (§\ref{sec:dual_execution}).

\paragraph{Hardware.} Each run is trained on 4$\times$NVIDIA A100 GPUs. The \gls{sql} and \gls{queryplan} runs are given an identical compute budget, so the comparison in §\ref{sec:claim3_training} is matched by construction.

\section{LLM-as-Judge Prompt}
\label{sec:llm_judge_prompt}

Error categorization uses Claude Haiku 4.5 at temperature~0.
The input always uses SQLite queries: for \gls{queryplan} models, the predicted plan is converted back to SQLite before classification.

\paragraph{Categories.}
\begin{itemize}
  \item \textbf{Schema Linking}: Wrong table or column name, hallucinated schema element.
  \item \textbf{Filtering / Logic}: Wrong WHERE predicate, JOIN condition, subquery logic, or conditional expression.
  \item \textbf{Aggregation / Grouping}: Wrong aggregation function, incorrect GROUP BY or HAVING clause.
  \item \textbf{Ordering / Ranking}: Wrong ORDER BY column or direction, wrong LIMIT/OFFSET, RANK vs.\ ROW\_NUMBER on ties; also covers structural variants for top-1 selection (e.g.\ MAX subquery vs.\ ORDER BY LIMIT~1).
  \item \textbf{Dialect Syntax}: Syntax error, dialect-specific function not valid in SQLite, malformed output.
  \item \textbf{Other}: Genuinely unclassifiable, or ambiguous question where multiple structurally incomparable SQL queries are defensible.
\end{itemize}

\paragraph{User prompt template.}
\begin{quote}
\small
\begin{verbatim}
Question: {question}

Gold SQL:
{gold_sql}

Predicted SQL:
{pred_sql}

Error message (if any): {error_msg}
\end{verbatim}
\end{quote}

The system prompt includes one few-shot example per category to anchor the classifier. The full system prompt is available in the released code.

\section{Datasets, Licenses, and Intended Use}
\label{sec:appendix_datasets}

We use three publicly released datasets, each within its intended research use.

\paragraph{BIRD~\cite{li2024can}.}
Released under CC BY-SA 4.0. We use the train split (9{,}428 question--SQL pairs over 69 databases) for fine-tuning and the dev split (1{,}534 questions over 11 databases) for evaluation. The hidden test set is not publicly available.

\paragraph{Spider~\cite{yu2018spider}.}
Released under CC BY-SA 4.0. We use the dev split (1{,}034 questions) for evaluation only. Spider is intended as a complex, cross-domain semantic-parsing and text-to-SQL benchmark; cross-dialect evaluation, which is what we use it for, is consistent with that purpose.

\paragraph{SynSQL-2.5M~\cite{li2025omnisql}.}
Released under Apache 2.0. The full dataset contains 2{,}544{,}390 \textit{(database, question, SQL, chain-of-thought)} quads over 16{,}583 synthetic databases. We sample 1M examples from this corpus, mixed with \texttt{BIRD-train}, as the fine-tuning source described in §\ref{sec:training}. We do not redistribute or modify the underlying data; the full corpus remains available from its original release.

All three datasets are English-only and target SQLite at source. Our cross-dialect evaluation derives PostgreSQL, MySQL, and ClickHouse copies via schema-level ETL from the released SQLite databases (§\ref{sec:dual_execution}); the natural-language questions and gold queries are unchanged.

\section{\metricname{} Validation Details}
\label{sec:appendix_metric_validation}

We validate \metricname{} on three fronts: the accuracy of the question classifier that drives its per-class comparison strategy, agreement with human judgment on cases where \metricname{} disagrees with \gls{birdea}, and the effect of the metric choice on the reported reliability of the conversion pipeline itself.

\paragraph{Classifier validation.}
\metricname{}'s comparison strategy hinges on correctly classifying each question along the ordering and row-count axes (§\ref{sec:multimetric:classification}). We validate the LLM classifier on 103 BIRD questions annotated by hand along both axes, reporting balanced accuracy per axis: \textbf{93.44\%} on ordering sensitivity and \textbf{99.50\%} on row-count sensitivity. The two axes are decided independently. Questions misclassified as \texttt{not\_ordered} fall back to \gls{birdea}-like set comparison, which is strictly more permissive than the ordered variants and is therefore the dominant residual source of false positives in \metricname{} (see Appendix~\ref{sec:metric_conservatism}); the empirical false-positive rate is bounded by the classifier's per-axis accuracy reported above.

\paragraph{Human agreement on disagreement cases.}
We sampled 103 \texttt{BIRD} predictions on which \metricname{} and \gls{birdea} reach opposite verdicts, and had a human annotator decide which verdict matches the intended answer to the natural-language question. Two annotators contributed, both authors with strong \gls{sql} proficiency, with each example labelled by exactly one of them. The instructions target semantic correctness rather than raw result equality: they state that different \gls{sql} queries can produce the same correct answer, and that differing row order where the question does not require one, differing column names, and harmless extra columns should all be accepted. As a control against favouring our own representation, annotators never saw the \gls{queryplan}: every prediction was compiled to SQLite before display, so labellers were blind to both the generation target and the dialect it came from. Table~\ref{tab:human_agreement} shows that humans side with \metricname{} in 96.1\% of cases and with \gls{birdea} in 3.9\%. A McNemar test on the contingency table gives $\chi^2=85.79$, $p\approx 2\times 10^{-20}$, effect size 0.92 (large). Concretely, on the set where the two metrics disagree, the disagreement is a \gls{birdea} misclassification roughly 25$\times$ more often than a \metricname{} misclassification. We note two limitations of this study: annotations are single-labeler per example (no inter-annotator agreement reported), and the sample is drawn from the disagreement set rather than i.i.d.\ from \texttt{BIRD}.

\begin{table}[t]
  \centering
  \small
  \begin{tabular}{l r r}
    \toprule
    \textbf{Metric} & \textbf{Count} & \textbf{\%} \\
    \midrule
    \metricname{}   & 99 & 96.1 \\
    \gls{birdea}    & \phantom{0}4 & \phantom{0}3.9 \\
    \bottomrule
  \end{tabular}
  \caption{Human agreement on 103 \texttt{BIRD} cases where \metricname{} and \gls{birdea} disagree. McNemar $\chi^2=85.79$, $p=2\times 10^{-20}$.}
  \label{tab:human_agreement}
\end{table}

\paragraph{Annotator instructions.}
The annotator received the instructions reproduced below alongside a per-example interface displaying the natural-language question, the gold SQL (left), the predicted SQL (right), and both result tables. Each example required one of three correctness verdicts (\textsc{Correct}, \textsc{Unsure}, \textsc{Incorrect}), with two optional flags for downstream filtering (\textsc{Gold-Is-Wrong}, \textsc{Other-Issue}) and an optional free-text note required only when a flag was set.

\textbf{Task.} Evaluate whether the predicted SQL correctly answers the natural-language question by comparing its result to the gold SQL's result. Focus on \emph{semantic correctness}: different SQL queries can produce the same correct answer.

\textbf{Verdicts.}
\begin{itemize}
  \item \textsc{Semantically Correct.} The predicted result answers the question correctly, even if the SQL itself differs. Same data in both result tables; same values, possibly in a different order if the question does not specify one; column names may differ; harmless extra columns are tolerated.
  \item \textsc{Unsure / Partially Correct.} Use sparingly, only for rare edge cases: results are correct but the logic to reach them is wrong, or correctness is genuinely uncertain. Most cases should resolve to \textsc{Correct} or \textsc{Incorrect}.
  \item \textsc{Incorrect.} The predicted SQL is clearly wrong: wrong logic, missing or extra rows, or wrong columns returned.
\end{itemize}

\textbf{Flags (optional).}
\begin{itemize}
  \item \textsc{Gold-Is-Wrong}: the gold reference itself appears incorrect.
  \item \textsc{Other-Issue}: ambiguous question, schema problem, or technical issue.
\end{itemize}

A free-text note field is shown but only required to be filled when a flag is set.

\paragraph{Pipeline reliability under both metrics.}
Table~\ref{tab:pipeline_reliability_both_metrics} re-runs the round-trip analysis of Section~\ref{sec:pipeline_reliability} on \texttt{BIRD} under \gls{birdea} alongside \metricname{}. Because the round-trip executes the \emph{same} logical query on the \emph{same} data in two different dialects, any ``failure'' under \gls{birdea} is either a genuine conversion defect or a metric artefact. The gap between the two rows per dialect is therefore a direct measurement of how much of the reported unreliability is metric-induced. Under \gls{birdea}, apparent reliability drops by 13--16 pp on every dialect including SQLite, with no corresponding change in the underlying conversion: the drop is explained by cross-dialect row-order nondeterminism in grouped results and by boundary-tie multiplicity that \gls{birdea} has no mechanism to tolerate. This result is a cautionary tale in its own right: any cross-dialect evaluation conducted under \gls{birdea} will substantially underreport the pipeline or model it is evaluating.

\begin{table}[t]
  \centering
  \small
  \setlength{\tabcolsep}{4pt}
  \begin{tabular}{l rrrr}
    \toprule
    \textbf{Metric} & \textbf{SQLite} & \textbf{PG} & \textbf{MySQL} & \textbf{CH} \\
    \midrule
    \gls{birdea}   & 83.1 & 81.2 & 79.7 & 80.6 \\
    \metricname{}  & 97.1 & 95.4 & 95.6 & 94.7 \\
    \midrule
    Gap & {\color{ForestGreen}+14.0} & {\color{ForestGreen}+14.2} & {\color{ForestGreen}+15.9} & {\color{ForestGreen}+14.1} \\
    \bottomrule
  \end{tabular}
  \caption{Round-trip conversion reliability (\%) on \texttt{BIRD} under \metricname{} vs.\ \gls{birdea}. PG = PostgreSQL, CH = ClickHouse. The consistent 14--16 pp gap per dialect is a metric artefact, not a pipeline defect.}
  \label{tab:pipeline_reliability_both_metrics}
\end{table}

\section{\metricname{} pseudocode}
\label{sec:metric_pseudocode}

The per-class dispatch of \metricname{} is summarized below. The full implementation, including column-mapping discovery, tie detection, and \texttt{LIMIT} boundary-alternative extraction, is released as part of the code accompanying this paper.

\begin{quote}
\footnotesize
\begin{verbatim}
function DF-MATCH(pred, gold, class):
    pred <- MAP-COLUMNS(pred, gold)
    if pred is None:  # gold column unmappable
        return false
    if class == "not_ordered":
        return MULTISET-EQUAL(pred, gold)
    if class == "ordered_fixed":
        if |pred| != |gold|:
            return false
        ties <- DETECT-TIES(gold.order_keys)
        return POSITIONAL-MATCH(
            pred, gold, tie_groups=ties)
    # class == "ordered_not_fixed"
    if |pred| < |gold|:
        return false
    pred <- pred[:|gold|]
    alts <- BOUNDARY-ALTERNATIVES(
        gold.order_keys, gold.limit)
    return ORDERED-MATCH(pred, gold,
        boundary_alts=alts)
\end{verbatim}
\end{quote}

\texttt{POSITIONAL-MATCH} accepts any permutation of rows that share a tie group (rows whose \texttt{ORDER BY} keys are equal up to $\varepsilon=10^{-9}$). \texttt{ORDERED-MATCH} enforces exact ordering for rows before the \texttt{LIMIT} cut and accepts any row at or beyond the cut whose key values match the pre-computed boundary alternatives. All numeric comparisons are $\varepsilon$-tolerant.

\section{\metricname{} conservatism: proof sketch}
\label{sec:metric_conservatism}

We sketch why \metricname{} produces no false positives, given a correct question classification (§\ref{sec:multimetric:classification}). A false positive is a verdict of ``correct'' for a predicted dataframe $\hat{T}$ that does not match the gold $T^*$ under the question's intended semantics. We argue per class.

\paragraph{\texttt{not\_ordered}.}
Comparison reduces to a multiset (or set, under \texttt{DISTINCT}-tolerance) equality on rows after column mapping. Column mapping requires every gold column to be matched to some predicted column with the same distinct values; an unmappable gold column rejects $\hat{T}$ outright. Given a valid mapping, the multiset/set check accepts only when every gold row has a matching predicted row (and, in the strict variant, vice versa). $\hat{T}$ cannot contain a row absent from $T^*$ without breaking the equality, so the verdict cannot be ``correct'' on a mismatched answer.

\paragraph{\texttt{ordered\_fixed}.}
Cardinality must match exactly, and rows are compared positionally, with permutations permitted only \emph{within tie groups}. Tie groups are computed from the gold ordering keys (Section~\ref{sec:multimetric:ties}): two rows share a tie group iff their gold-side \texttt{ORDER BY} expressions are equal (up to $\varepsilon$). A predicted row with a different ordering-key value cannot fall into the same tie group as the gold row at that position, and so cannot be permuted into place. Any divergence in the answer set or in the order of distinct-key rows is rejected.

\paragraph{\texttt{ordered\_not\_fixed}.}
Predicted output is truncated to $|T^*|$ rows; if $|\hat{T}| < |T^*|$, the verdict is ``incorrect''. The truncated prefix is then compared as in \texttt{ordered\_fixed}, with one extension: a row at the \texttt{LIMIT} boundary may be replaced by a row from the boundary-alternatives query, which enumerates exactly those rows the gold query \emph{could have legitimately returned} at the cut (Section~\ref{sec:multimetric:ties}). By construction, the alternatives set contains only rows whose ordering-key values equal the gold row's at the boundary, so accepting one in place of the gold row preserves the gold query's semantics.

\paragraph{Conservatism caveat.}
The argument assumes a correct question classification and a faithfully extractable order-key/boundary-alternatives query from the gold SQL. Misclassification of an ordered question as \texttt{not\_ordered} weakens the comparison to a set check (strictly more permissive), which can introduce false positives; this is the dominant residual failure mode and is bounded empirically by the classifier accuracy reported in §\ref{sec:metric_validation}.

\section{Feature-level meta-analysis: details}
\label{sec:appendix_features}

This appendix provides the target definition, model list, feature definitions, pooling protocol, and full results behind §\ref{sec:claim4_structural}.

\paragraph{Target.}
We define \verb|qp_beats_sql| on \emph{divergent} queries, those where exactly one of \gls{sql} and \gls{queryplan} is correct, with \texttt{qp\_beats\_sql}=1 iff the \gls{queryplan} is the correct one. Restricting to divergent rows isolates the representation effect from query difficulty: queries where both sides agree (both correct, or both wrong) carry no signal about which representation a model prefers. As a sanity check, a difficulty-only baseline (predicting per-model task accuracy from the same features) achieves cross-validated AUC 0.50--0.65, well below the 0.74--0.87 the same features achieve when targeting individual \gls{sql} and \gls{queryplan} outcomes; \verb|qp_beats_sql| is therefore not just a re-encoded difficulty target.

\paragraph{Models.} The meta-analysis pools per-model coefficients across 13 models spanning 6 families and four size tiers:
Ministral-3-3B,
Ministral-3-8B (prompted and LoRA fine-tuned on \gls{sql}),
Gemma 4 E4B,
Qwen 3.5 9B,
GPT-OSS 20B,
Qwen 3.5 27B,
Gemma 4 31B,
Qwen 3.5 122B,
Mistral Medium,
GPT-OSS 120B,
Claude Sonnet 4.6,
GPT-5.5,
and Gemini 3.1 Pro.

\paragraph{Features.} Twenty-one query-level structural features enter the elastic-net regression, organized into four conceptual clusters (\textbf{A}--\textbf{D}; 17 features) plus two derived families (\texttt{ps\_*}, \texttt{ax\_*}; 4 features). Names use the prefix conventions \texttt{sc\_*} (\gls{sql} complexity from a sqlglot parse), \texttt{ps\_*} (plan structure from the gold plan tree), \texttt{qc\_*} (categorizations of either side), and \texttt{ax\_*} (cross-representation asymmetry).

\emph{Cluster A, Planning horizon} (how far across the \gls{sql} text dependent information appears):
\verb|qc_scope_cross_clause| (column alias used in \texttt{ORDER BY}/\texttt{GROUP BY}/\texttt{HAVING}, or positional \texttt{ORDER BY});
\verb|qc_n_forward_refs| (count of identifier uses appearing textually before their definition);
\verb|qc_max_alias_distance| (largest character distance between a forward use and its definition);
\verb|qc_type_coercion_level| (0--3 ordinal, combining cast counts and date-function presence);
\verb|qc_n_casts| (\texttt{CAST}/\texttt{TRY\_CAST} count);
\verb|qc_has_date_functions| (any of \texttt{strftime}, \texttt{date}, \texttt{format\_date}, \texttt{to\_char}, etc.);
\verb|qc_has_implicit_coercion| (numeric-vs-string comparison in a predicate).

\emph{Cluster B, Relational topology} (shape of the gold plan tree):
\verb|qc_dag_max_width| (max number of nodes at any depth);
\verb|qc_max_join_depth| (longest \textsc{Join}$\to$\textsc{Join} chain).

\emph{Cluster C, Semantic complexity} (what the query computes):
\verb|qc_n_reasoning_steps| (non-leaf plan node count);
\verb|qc_literal_diversity_level| (0--3 ordinal over literal count, type diversity, date and pattern presence);
\verb|qc_n_literals|;
\verb|qc_has_pattern_literal| (string literal containing \texttt{\%} or \texttt{\_});
\verb|qc_category_ordered_fixed| and \verb|qc_category_ordered_not_fixed| (two one-hot indicators expanding the result-shape contract from §\ref{sec:multimetric:classification}; \texttt{not\_ordered} is the reference category).

\emph{Cluster D, Format gap} (structural differences between \gls{sql} and plan):
\verb|qc_locality_gap| ($\text{plan locality} - \text{\gls{sql} locality}$, where locality is the fraction of identifier uses within a short distance of their definition);
\verb|qc_n_ambiguous_refs| (count of unqualified column references in multi-table queries).

\emph{Plan structure (\texttt{ps\_*})}:
\verb|ps_max_depth| (longest root-to-leaf path);
\verb|ps_has_scalar_subquery| (any plan node with a nested scalar-subquery operator).

\emph{Asymmetry (\texttt{ax\_*})}:
\verb|ax_length_ratio| ($\texttt{ps\_n\_nodes} / \max(\texttt{sc\_n\_tokens},1)$);
\verb|ax_length_diff| ($\texttt{ps\_n\_nodes} - \texttt{sc\_n\_tokens}$).

The full \texttt{sc\_*} family is computed but excluded from the regression: its signal is captured by the \texttt{ps\_*} and \texttt{qc\_*} plan-side features plus the empirical difficulty confounder.

\paragraph{Pooling.} For each model and each binary target we fit an elastic-net logistic regression on the 21 features above with bootstrap resampling. Per-model coefficients $\hat\beta_m$ and standard errors $\text{se}_m$ are pooled via inverse-variance-weighted fixed-effect meta-analysis: $\hat\beta = \sum_m w_m \hat\beta_m / \sum_m w_m$ with $w_m = 1/\text{se}_m^2$. We report Cochran's $I^2$ as the heterogeneity diagnostic, the proportion of total variation in $\hat\beta_m$ attributable to between-model differences rather than sampling error. High $I^2$ indicates a feature whose effect is model-specific and pools to near-zero by cancellation; low $I^2$ with a non-zero pooled estimate indicates a universal effect.

\paragraph{Results.}
Of 21 features, only one has a pooled CI excluding zero: \verb|ps_has_scalar_subquery| ($\hat\beta=-0.04$, 95\% CI $[-0.07,-0.004]$, $I^2=66\%$). All others average to near-zero with inconsistent per-model signs. Per-model preferences do exist but disagree across models (e.g., Claude Sonnet 4.6 favors \gls{sql} on ordered queries, while GPT-OSS 120B favors \glspl{queryplan} with scalar subqueries), consistent with model-internal factors such as tokenization and format familiarity that query-level features cannot capture.

\section{Inference Overhead and Compiler Extension Cost}
\label{sec:appendix_overhead}

\paragraph{Token cost of the representation.}
\Glspl{queryplan} are more verbose than the \gls{sql} they replace. Measured over \texttt{BIRD}, the emitted plan averages 227 output tokens against 58 for the corresponding \gls{sql} query, so the representation itself costs about $3.9\times$ the tokens of the query it replaces.

This artefact cost is not the whole generation cost. For both targets, most output tokens are reasoning rather than the query artefact: of roughly 319 output tokens per \gls{sql} generation only about 58 are the query, and of roughly 1685 per \gls{queryplan} generation only about 227 are the plan. End to end the ratio is therefore larger than $3.9\times$, because both formats carry a long shared reasoning trace that is not specific to either representation. Reducing reasoning verbosity shrinks the absolute gap for both targets, since the representation accounts for an absolute difference of roughly 170 tokens.

We report output tokens rather than wall-clock latency. Tokens are the hardware-independent quantity; latency depends on serving stack, batch size, and hardware, and we did not measure it under controlled conditions.

\paragraph{Cost of supporting a new dialect.}
Adding a backend takes roughly one engineer-day, because we do not implement a dialect from scratch. Calcite ships approximately 40 dialects, including Snowflake and BigQuery, and the work is to align their semantics with the conventions our pipeline assumes: NULL ordering, date handling, cast behaviour, and identifier case sensitivity (§\ref{sec:pipeline}). In our experience the larger cost was not the dialect but the benchmark environment, that is migrating and hosting four parallel databases with identical contents. That is a cost of \emph{measuring} \gls{portability}, not of deploying it: a production system targets one backend it already runs.

\end{document}

%% file: figures/fig_portability_table.tex
\begin{table*}[t]
\centering
\small
\setlength{\tabcolsep}{3pt}
\begin{tabular}{ll rrrrr rrrrr}
\toprule
 & & \multicolumn{5}{c}{\textbf{BIRD-dev}} & \multicolumn{5}{c}{\textbf{Spider-dev}} \\
\cmidrule(lr){3-7} \cmidrule(lr){8-12}
\textbf{Model} & \textbf{Target} & \textbf{SQLite} & \textbf{PG} & \textbf{MySQL} & \textbf{CH} & \textbf{Port.} & \textbf{SQLite} & \textbf{PG} & \textbf{MySQL} & \textbf{CH} & \textbf{Port.} \\
\midrule
\multicolumn{12}{l}{\textit{Small}} \\
Ministral 3-8B$^\dagger$ & SQL & 61.2 & 52.4 & 57.4 & 50.3 & 82.1 & 85.3 & 78.3 & 81.5 & 77.6 & 90.9 \\
Ministral 3-8B$^\dagger$ & QP & 64.5 & 64.2 & 65.2 & 63.5 & 97.3 & 84.1 & 83.2 & 84.5 & 82.9 & 98.2 \\
SQLCoder & SQL & 39.1 & 29.3 & 35.1 & 27.6 & 70.6 & 66.0 & 18.4 & 50.3 & 18.2 & 27.5 \\
OmniSQL & SQL & 62.1 & 54.9 & 58.0 & 48.7 & 78.5 & 86.0 & 76.1 & 83.0 & 76.5 & 88.5 \\
Qwen 3.5 9B & SQL & 66.6 & 58.7 & 62.5 & 55.5 & 83.4 & 86.4 & 79.1 & 81.9 & 78.0 & 90.2 \\
Qwen 3.5 9B & QP & 26.8 & 26.5 & 26.9 & 26.7 & 98.3 & 53.1 & 52.4 & 52.9 & 54.4 & 96.4 \\
Gemma 4 E2B & SQL & 46.4 & 41.9 & 42.7 & 38.2 & 82.3 & 74.8 & 67.9 & 69.9 & 68.9 & 90.7 \\
Gemma 4 E2B & QP & 10.0 & 10.0 & 9.9 & 9.9 & 99.4 & 23.3 & 23.1 & 23.4 & 23.2 & 98.8 \\
Gemma 4 E4B & SQL & 58.8 & 51.9 & 52.4 & 47.3 & 80.4 & 83.2 & 72.3 & 75.7 & 73.4 & 86.9 \\
Gemma 4 E4B & QP & 27.4 & 27.6 & 27.7 & 27.8 & 98.6 & 53.9 & 53.2 & 53.4 & 53.4 & 98.7 \\
Ministral 3-3B & SQL & 43.4 & 39.8 & 43.2 & 34.9 & 80.6 & 71.0 & 66.1 & 69.7 & 64.8 & 91.2 \\
Ministral 3-3B & QP & 11.8 & 11.8 & 11.9 & 12.1 & 97.3 & 17.9 & 18.7 & 18.0 & 18.0 & 95.3 \\
Ministral 3-8B & SQL & 53.3 & 49.7 & 49.9 & 38.4 & 72.0 & 82.0 & 74.7 & 80.8 & 69.4 & 84.6 \\
Ministral 3-8B & QP & 15.7 & 16.1 & 16.1 & 16.0 & 97.2 & 33.8 & 36.3 & 33.1 & 33.6 & 91.2 \\
Ministral 3-14B & SQL & 54.4 & 51.4 & 52.9 & 37.4 & 68.7 & 82.0 & 77.6 & 81.0 & 69.5 & 84.7 \\
Ministral 3-14B & QP & 18.3 & 18.7 & 19.1 & 18.4 & 95.9 & 51.8 & 51.0 & 51.6 & 51.5 & 98.5 \\
\midrule
\multicolumn{12}{l}{\textit{Medium}} \\
GPT OSS 20B & SQL & 58.1 & 56.7 & 55.1 & 48.6 & 83.5 & 80.3 & 76.0 & 81.4 & 72.3 & 88.9 \\
GPT OSS 20B & QP & 48.5 & 48.8 & 49.4 & 48.9 & 98.3 & 73.6 & 72.6 & 73.4 & 73.1 & 98.7 \\
Qwen 3.5 27B & SQL & 69.0 & 65.1 & 67.0 & 59.9 & 86.9 & 87.7 & 83.0 & 80.0 & 82.0 & 91.3 \\
Qwen 3.5 27B & QP & 39.5 & 39.1 & 39.8 & 39.4 & 98.2 & 67.9 & 67.0 & 67.8 & 67.0 & 98.7 \\
Gemma 4 26B & SQL & 67.8 & 62.5 & 62.7 & 56.2 & 82.9 & 89.8 & 80.2 & 80.3 & 83.2 & 89.3 \\
Gemma 4 26B & QP & 54.6 & 54.3 & 54.8 & 54.6 & 99.1 & 83.3 & 82.3 & 83.1 & 82.2 & 98.7 \\
Gemma 4 31B & SQL & 70.6 & 64.8 & 66.8 & 58.9 & 83.4 & 88.9 & 75.7 & 78.4 & 81.6 & 85.2 \\
Gemma 4 31B & QP & 63.9 & 64.2 & 63.7 & 63.2 & 98.5 & 86.3 & 84.6 & 84.7 & 83.5 & 96.7 \\
\midrule
\multicolumn{12}{l}{\textit{Large}} \\
GPT OSS 120B & SQL & 64.6 & 61.1 & 61.3 & 55.1 & 85.3 & 84.9 & 80.4 & 84.9 & 80.1 & 94.3 \\
GPT OSS 120B & QP & 58.7 & 58.6 & 59.8 & 58.2 & 97.4 & 79.8 & 79.1 & 80.1 & 78.8 & 98.4 \\
Mistral Medium 35 & SQL & 63.1 & 59.2 & 60.1 & 51.8 & 82.0 & 84.7 & 80.9 & 84.7 & 81.3 & 95.5 \\
Mistral Medium 35 & QP & 55.9 & 55.5 & 56.1 & 55.3 & 98.5 & 78.0 & 76.7 & 77.3 & 76.7 & 98.4 \\
Claude Sonnet 4.6 & SQL & 69.4 & 65.8 & 67.2 & 60.1 & 86.6 & 87.0 & 86.1 & 86.6 & 85.1 & 97.9 \\
Claude Sonnet 4.6 & QP & 64.9 & 64.3 & 64.9 & 63.9 & 98.4 & 84.9 & 83.4 & 84.1 & 83.1 & 97.9 \\
GPT 5.5 & SQL & 67.3 & 65.8 & 65.2 & 58.0 & 86.1 & 85.3 & 84.0 & 85.7 & 85.0 & 98.0 \\
GPT 5.5 & QP & 64.9 & 64.1 & 63.8 & 62.6 & 96.4 & 85.6 & 84.9 & 84.9 & 83.7 & 97.7 \\
Gemini 3.1 Pro & SQL & \textbf{73.2} & \textbf{69.6} & \textbf{70.0} & 63.1 & 86.2 & \textbf{91.9} & 79.3 & 81.6 & 88.5 & 86.3 \\
Gemini 3.1 Pro & QP & 68.2 & 67.5 & 68.8 & \textbf{67.1} & 97.4 & 90.4 & \textbf{89.2} & \textbf{90.1} & \textbf{88.9} & 98.4 \\
\bottomrule
\end{tabular}
\caption{Per-dialect execution accuracy (\%) on BIRD-dev and Spider-dev under \metricname{}; all prompted runs use greedy decoding (single sample). PG = PostgreSQL, CH = ClickHouse. Port.\ = portability, defined as worst-dialect / best-dialect accuracy (higher = more portable; 100.0 = identical across dialects). $^\dagger$ LoRA fine-tuned.}
\label{tab:portability_results}
\end{table*}

%% file: sample.bib
@inproceedings{ananthakrishnan2026querygym,
  title={QueryGym: Step-by-Step Interaction with Relational Databases},
  author={Ananthakrishnan, Haritha and Kokel, Harsha and Sikes, Kelsey and Bhattacharjya, Debarun and Katz, Michael and Sohrabi, Shirin and Srinivas, Kavitha},
  booktitle={Proceedings of the AAAI Conference on Artificial Intelligence},
  volume={40},
  pages={41544--41546},
  year={2026}
}

@inproceedings{zelle1996learning,
  title={Learning to parse database queries using inductive logic programming},
  author={Zelle, John M and Mooney, Raymond J},
  booktitle={Proceedings of the national conference on artificial intelligence},
  pages={1050--1055},
  year={1996}
}

@inproceedings{wang2020rat,
  title={Rat-sql: Relation-aware schema encoding and linking for text-to-sql parsers},
  author={Wang, Bailin and Shin, Richard and Liu, Xiaodong and Polozov, Oleksandr and Richardson, Matthew},
  booktitle={Proceedings of the 58th annual meeting of the association for computational linguistics},
  pages={7567--7578},
  year={2020}
}

@article{li2025omnisql,
  title={Omnisql: Synthesizing high-quality text-to-sql data at scale},
  author={Li, Haoyang and Wu, Shang and Zhang, Xiaokang and Huang, Xinmei and Zhang, Jing and Jiang, Fuxin and Wang, Shuai and Zhang, Tieying and Chen, Jianjun and Shi, Rui and others},
  journal={arXiv preprint arXiv:2503.02240},
  year={2025}
}

@misc{substrait,
  author = {substrait-io},
  title = {Substrait: Cross-Language Serialization for Relational Algebra},
  year = {2021},
  month = {8},
  day = {31},
  publisher = {GitHub},
  journal = {GitHub repository},
  howpublished = {\url{https://github.com/substrait-io/substrait}}
}

@article{li2024can,
  title={Can llm already serve as a database interface? a big bench for large-scale database grounded text-to-sqls},
  author={Li, Jinyang and Hui, Binyuan and Qu, Ge and Yang, Jiaxi and Li, Binhua and Li, Bowen and Wang, Bailin and Qin, Bowen and Geng, Ruiying and Huo, Nan and others},
  journal={Advances in Neural Information Processing Systems},
  volume={36},
  year={2024}
}

@misc{Perlitz2026PolySQLST,
  title={PolySQL: Scaling Text-to-SQL Evaluation Across SQL Dialects via Automated Backend Isomorphism},
  author={Yotam Perlitz and Elad Venezian and Corentin Royer and Francesco Fusco and Andrea Giovannini},
  howpublished={arXiv preprint arXiv:2605.07796},
  year={2026},
  url={https://arxiv.org/abs/2605.07796}
}

@article{gao2024dailsql,
  title={Text-to-SQL empowered by large language models: A benchmark evaluation},
  author={Gao, Dawei and Wang, Haibin and Li, Yaliang and Sun, Xiuyu and Qian, Yichen and Ding, Bolin and Zhou, Jingren},
  journal={Proceedings of the VLDB Endowment},
  volume={17},
  number={5},
  pages={1132--1145},
  year={2024}
}

@article{pourreza2024dinsql,
  title={DIN-SQL: Decomposed in-context learning of text-to-SQL with self-correction},
  author={Pourreza, Mohammadreza and Rafiei, Davood},
  journal={Advances in Neural Information Processing Systems},
  volume={36},
  year={2024}
}

@article{talaei2024chess,
  title={CHESS: Contextual harnessing for efficient SQL synthesis},
  author={Talaei, Shayan and Pourreza, Mohammadreza and Chang, Yu-Chen and Mirhoseini, Azalia and Saberi, Amin},
  journal={arXiv preprint arXiv:2405.16755},
  year={2024}
}

@inproceedings{yu2018spider,
  title={Spider: A large-scale human-labeled dataset for complex and cross-domain semantic parsing and text-to-sql task},
  author={Yu, Tao and Zhang, Rui and Yang, Kai and Yasunaga, Michihiro and Wang, Dongxu and Li, Zifan and Ma, James and Li, Irene and Yao, Qingning and Roman, Shanelle and others},
  booktitle={Proceedings of the 2018 conference on empirical methods in natural language processing},
  pages={3911--3921},
  year={2018}
}

@misc{sqlglot,
  title={SQLGlot: A Python SQL parser, transpiler, optimizer, and engine},
  author={Mao, Toby},
  year={2023},
  howpublished={\url{https://github.com/tobymao/sqlglot}}
}

@misc{calcite,
  title={Apache Calcite: A dynamic data management framework},
  author={{Apache Software Foundation}},
  year={2024},
  howpublished={\url{https://calcite.apache.org/}}
}

@article{begoli2018apache,
  title={Apache Calcite: A foundational framework for optimized query processing over heterogeneous data sources},
  author={Begoli, Edmon and Camacho-Rodr{\'\i}guez, Jes{\'u}s and Hyde, Julian and Mior, Michael J and Lemire, Daniel},
  journal={Proceedings of the 2018 International Conference on Management of Data},
  pages={221--230},
  year={2018}
}

@inproceedings{guo2019irnet,
  title={Towards Complex Text-to-{SQL} in Cross-Domain Database with Intermediate Representation},
  author={Guo, Jiaqi and Zhan, Zecheng and Gao, Yan and Xiao, Yan and Lou, Jian-Guang and Liu, Ting and Zhang, Dongmei},
  booktitle={Proceedings of the 57th Annual Meeting of the Association for Computational Linguistics},
  pages={4524--4535},
  year={2019}
}

@misc{eyal2023qpl,
  title={Semantic Parsing for Complex Data Retrieval: Targeting Query Plans vs.\ {SQL} for No-Code Access to Relational Databases},
  author={Eyal, Ben and Bachar, Amir and Haroche, Ophir and Elhadad, Michael},
  howpublished={arXiv preprint arXiv:2312.14798},
  year={2023}
}

@misc{zhang2026dial,
  title={Dial: A Knowledge-Grounded Dialect-Specific {NL2SQL} System},
  author={Zhang, Xiang and Xu, Hongming and Zhou, Le and Zhou, Wei and Zhou, Xuanhe and Li, Guoliang and Luo, Yuyu and Liu, Changdong and Chen, Guorun and Liao, Jiang and Wu, Fan},
  howpublished={arXiv preprint arXiv:2603.07449},
  year={2026}
}

@inproceedings{zhong2020testsuite,
  title={Semantic Evaluation for Text-to-{SQL} with Distilled Test Suites},
  author={Zhong, Ruiqi and Yu, Tao and Klein, Dan},
  booktitle={Proceedings of the 2020 Conference on Empirical Methods in Natural Language Processing},
  pages={396--411},
  year={2020}
}

@misc{ascoli2024etm,
  title={{ETM}: Modern Insights into Perspective on Text-to-{SQL} Evaluation in the Age of Large Language Models},
  author={Ascoli, Benjamin G. and Kandikonda, Yasoda Sai Ram and Choi, Jinho D.},
  howpublished={arXiv preprint arXiv:2407.07313},
  year={2024}
}

@misc{kim2024flex,
  title={{FLEX}: Expert-level False-Less {EX}ecution Metric for Reliable Text-to-{SQL} Benchmark},
  author={Kim, Heegyu and Jeon, Taeyang and Choi, Seunghwan and Choi, Seungtaek and Cho, Hyunsouk},
  howpublished={arXiv preprint arXiv:2409.19014},
  year={2024}
}

@misc{dlt,
  title = {dlt: the open-source {Python} library for data loading},
  author = {{dltHub}},
  year = {2022},
  howpublished = {\url{https://github.com/dlt-hub/dlt}},
  note = {Accessed 2026-05-11}
}

@inproceedings{hu2022lora,
  title={{LoRA}: Low-rank adaptation of large language models},
  author={Hu, Edward J and Shen, Yelong and Wallis, Phillip and Allen-Zhu, Zeyuan and Li, Yuanzhi and Wang, Shean and Wang, Liang and Chen, Weizhu},
  booktitle={International Conference on Learning Representations (ICLR)},
  year={2022}
}

@article{li2024codes,
  title={Codes: Towards building open-source language models for text-to-sql},
  author={Li, Haoyang and Zhang, Jing and Liu, Hanbing and Fan, Ju and Zhang, Xiaokang and Zhu, Jun and Wei, Renjie and Pan, Hongyan and Li, Cuiping and Chen, Hong},
  journal={Proceedings of the ACM on Management of Data},
  volume={2},
  number={3},
  pages={1--28},
  year={2024},
  publisher={ACM New York, NY, USA}
}

@inproceedings{lei2025spider,
  title={Spider 2.0: Evaluating language models on real-world enterprise text-to-sql workflows},
  author={Lei, Fangyu and Chen, Jixuan and Ye, Yuxiao and Cao, Ruisheng and Shin, Dongchan and Su, Hongjin and Suo, Zhaoqing and Gao, Hongcheng and Hu, Wenjing and Yin, Pengcheng and others},
  booktitle={International Conference on Learning Representations},
  year={2025}
}

@article{pourreza2024sql,
  title={Sql-gen: Bridging the dialect gap for text-to-sql via synthetic data and model merging},
  author={Pourreza, Mohammadreza and Sun, Ruoxi and Li, Hailong and Miculicich, Lesly and Pfister, Tomas and Arik, Sercan O},
  journal={arXiv preprint arXiv:2408.12733},
  year={2024}
}

@article{lin2024momq,
  title={Momq: Mixture-of-experts enhances multi-dialect query generation across relational and non-relational databases},
  author={Lin, Zhisheng and Liu, Yifu and Luo, Zhiling and Gao, Jinyang and Li, Yu},
  journal={arXiv preprint arXiv:2410.18406},
  year={2024}
}

@inproceedings{scholak2021picard,
  title={PICARD: Parsing incrementally for constrained auto-regressive decoding from language models},
  author={Scholak, Torsten and Schucher, Nathan and Bahdanau, Dzmitry},
  booktitle={Proceedings of the 2021 conference on empirical methods in natural language processing},
  pages={9895--9901},
  year={2021}
}

@inproceedings{geng2023grammar,
  title={Grammar-constrained decoding for structured NLP tasks without finetuning},
  author={Geng, Saibo and Josifoski, Martin and Peyrard, Maxime and West, Robert},
  booktitle={Proceedings of the 2023 Conference on Empirical Methods in Natural Language Processing},
  pages={10932--10952},
  year={2023}
}

@inproceedings{dong2016language,
  title={Language to logical form with neural attention},
  author={Dong, Li and Lapata, Mirella},
  booktitle={Proceedings of the 54th Annual Meeting of the Association for Computational Linguistics (Volume 1: Long Papers)},
  pages={33--43},
  year={2016}
}

@inproceedings{rabinovich2017asn,
  title={Abstract syntax networks for code generation and semantic parsing},
  author={Rabinovich, Maxim and Stern, Mitchell and Klein, Dan},
  booktitle={Proceedings of the 55th Annual Meeting of the Association for Computational Linguistics (Volume 1: Long Papers)},
  pages={1139--1149},
  year={2017}
}

@inproceedings{yin2018tranx,
  title={TRANX: A transition-based neural abstract syntax parser for semantic parsing and code generation},
  author={Yin, Pengcheng and Neubig, Graham},
  booktitle={Proceedings of the 2018 conference on empirical methods in natural language processing: System demonstrations},
  pages={7--12},
  year={2018}
}

@article{zheng2023judging,
  title={Judging llm-as-a-judge with mt-bench and chatbot arena},
  author={Zheng, Lianmin and Chiang, Wei-Lin and Sheng, Ying and Zhuang, Siyuan and Wu, Zhanghao and Zhuang, Yonghao and Lin, Zi and Li, Zhuohan and Li, Dacheng and Xing, Eric and others},
  journal={Advances in neural information processing systems},
  volume={36},
  pages={46595--46623},
  year={2023}
}

@article{panickssery2024llm,
  title={Llm evaluators recognize and favor their own generations},
  author={Panickssery, Arjun and Bowman, Samuel R and Feng, Shi},
  journal={Advances in Neural Information Processing Systems},
  volume={37},
  pages={68772--68802},
  year={2024}
}

@inproceedings{zhang2025exesql,
  title={ExeSQL: Self-Taught Text-to-SQL Models with Execution-Driven Bootstrapping for SQL Dialects},
  author={Zhang, Jipeng and Yang, Haolin and Miao, Kehao and Zhang, Ruiyuan and Pi, Renjie and Gao, Jiahui and Zhou, Xiaofang},
  booktitle={Findings of the Association for Computational Linguistics: EMNLP 2025},
  pages={24305--24326},
  year={2025}
}

@inproceedings{shi2025dialect,
  title={Dialect-SQL: An Adaptive Framework for Bridging the Dialect Gap in Text-to-SQL},
  author={Shi, Jie and Cao, Xi and Xu, Bo and Liang, Jiaqing and Xiao, Yanghua and Chen, Jia and Wang, Peng and Wang, Wei},
  booktitle={Proceedings of the 2025 Conference on Empirical Methods in Natural Language Processing},
  pages={3604--3619},
  year={2025}
}
